%% file: root.tex
\documentclass[letterpaper, 10 pt, conference]{ieeeconf}  % Comment this line out if you need a4paper

\IEEEoverridecommandlockouts                              % This command is only needed if 
\usepackage[resetlabels]{multibib}
\usepackage{subfiles}
\usepackage{csquotes}
\usepackage{gensymb}
\usepackage{pifont}
\usepackage{textcomp}
\usepackage[table]{xcolor}
\usepackage{todonotes}
\usepackage{blindtext}
\usepackage{soul}
\usepackage{graphicx}
\usepackage{mathtools}
\usepackage{multirow}
\usepackage{flushend}
\usepackage{booktabs}
\usepackage{array}
\usepackage{tabularx}

\usepackage{amsmath} 
\usepackage{amssymb} 
\usepackage{amsfonts}
\usepackage{bm} 
\usepackage{siunitx}
\usepackage{cancel}
\let\labelindent\relax
\usepackage{enumitem}
\usepackage{microtype}
\usepackage{xcolor}
\usepackage[breaklinks,colorlinks]{hyperref}
\usepackage{cite}
\usepackage[export]{adjustbox}
\usepackage[hang,flushmargin]{footmisc}
\usepackage{threeparttable}
\usepackage{cuted}
\newcites{S}{References}

\usepackage[capitalize]{cleveref}
\crefname{section}{Sec.}{Secs.}
\Crefname{section}{Section}{Sections}
\Crefname{table}{Table}{Tables}
\crefname{table}{Tab.}{Tabs.}

\definecolor{mygreen}{HTML}{00A64F}
\definecolor{myred}{HTML}{ED1B23}

\newcommand{\secref}[1]{Sec.~\ref{#1}}
\renewcommand{\eqref}[1]{Eq.~(\ref{#1})}
\newcommand{\figref}[1]{Fig.~\ref{#1}}
\newcommand{\tabref}[1]{Tab.~\ref{#1}}

\newcommand{\net}{ULOPS}
\newcommand{\la}{Uniform Evidence Loss}
\newcommand{\lc}{Contrastive Uncertainty Loss}
\newcommand{\lb}{Adaptive Uncertainty Separation Loss}

\renewcommand{\baselinestretch}{0.985}

\title{\LARGE \bf
Open-Set LiDAR Panoptic Segmentation \\ Guided by Uncertainty-Aware Learning
}

\author{Rohit Mohan$^{1}$, Julia Hindel$^{1}$, Florian Drews$^{2}$, Claudius Gläser$^{2}$, Daniele Cattaneo$^{1}$, Abhinav Valada$^{1}$% <-this % stops a space
\thanks{$^1$ Department of Computer Science, University of Freiburg, Germany}%
\thanks{$^2$ Bosch Research, Robert Bosch GmbH, Renningen, Germany}%
\thanks{This work was funded by the Bosch Research collaboration on AI-driven automated driving. Julia Hindel was supported by Deutsche Forschungsgemeinschaft (DFG, German Research Foundation) – SFB 1597 – 499552394.}% % <-this % stops a space
}

\begin{document}

\maketitle
\thispagestyle{empty}
\pagestyle{empty}

%%%%%%%%%%%%%%%%%%%%%%%%%%%%%%%%%%%%%%%%%%%%%%%%%%%%%%%%%%%%%%%%%%%%%%%%%%%%%%%%
\begin{abstract}
% Autonomous vehicles must navigate open-world environments and handle novel objects, yet existing LiDAR panoptic segmentation models operate under closed-set assumptions, failing to recognize unknown instances.
Autonomous vehicles that navigate in open-world environments may encounter previously unseen object classes. However, most existing LiDAR panoptic segmentation models rely on closed-set assumptions, failing to detect unknown object instances.
In this work, we propose \net, an uncertainty-guided open-set panoptic segmentation framework that leverages Dirichlet-based evidential learning to model predictive uncertainty. Our architecture incorporates separate decoders for semantic segmentation with uncertainty estimation, embedding with prototype association, and instance center prediction. During inference, we leverage uncertainty estimates to identify and segment unknown instances. To strengthen the model’s ability to differentiate between known and unknown objects, we introduce three uncertainty-driven loss functions. Uniform Evidence Loss to encourage high uncertainty in unknown regions. Adaptive Uncertainty Separation Loss ensures a consistent difference in uncertainty estimates between known and unknown objects at a global scale. Contrastive Uncertainty Loss refines this separation at the fine-grained level. To evaluate open-set performance, we extend benchmark settings on KITTI-360 and introduce a new open-set evaluation for nuScenes. Extensive experiments demonstrate that \net{} consistently outperforms existing open-set LiDAR panoptic segmentation methods. We make the code available at \url{http://ulops.cs.uni-freiburg.de}.
\end{abstract}

%%%%%%%%%%%%%%%%%%%%%%%%%%%%%%%%%%%%%%%%%%%%%%%%%%%%%%%%%%%%%%%%%%%%%%%%%%%%%%%%
\input{sections/01_introduction}

\input{sections/02_related-work}
\input{sections/03_methodology}

\input{sections/04_experiments}

\input{sections/05_conclusion}

{\footnotesize
\bibliographystyle{IEEEtran}
\bibliography{root}
}

\input{sections/supplementary}
\newpage

\end{document}

%% file: sections/01_introduction.tex
\section{Introduction}
Accurate scene understanding~\cite{mohan2023neural, luz2024amodal} plays a central role in autonomous driving, forming the foundation for safe navigation~\cite{vodisch2023codeps} and enhancing downstream tasks such as localization~\cite{cattaneo2025cmrnext, ballardini2021vehicle}. LiDAR 3D panoptic segmentation plays a crucial role in this task by detecting both semantic classes and individual object instances. However, models for LiDAR panoptic segmentation are typically trained on a closed-set of predefined categories~\cite{sirohi2021efficientlps}.
This constraint limits adaptability in open-world environments where novel objects frequently appear~\cite{hindel24topics, mohan2024panoptic}.

\begin{figure}
    \vspace{0.1em}
    \centering
    \includegraphics[width=\linewidth]{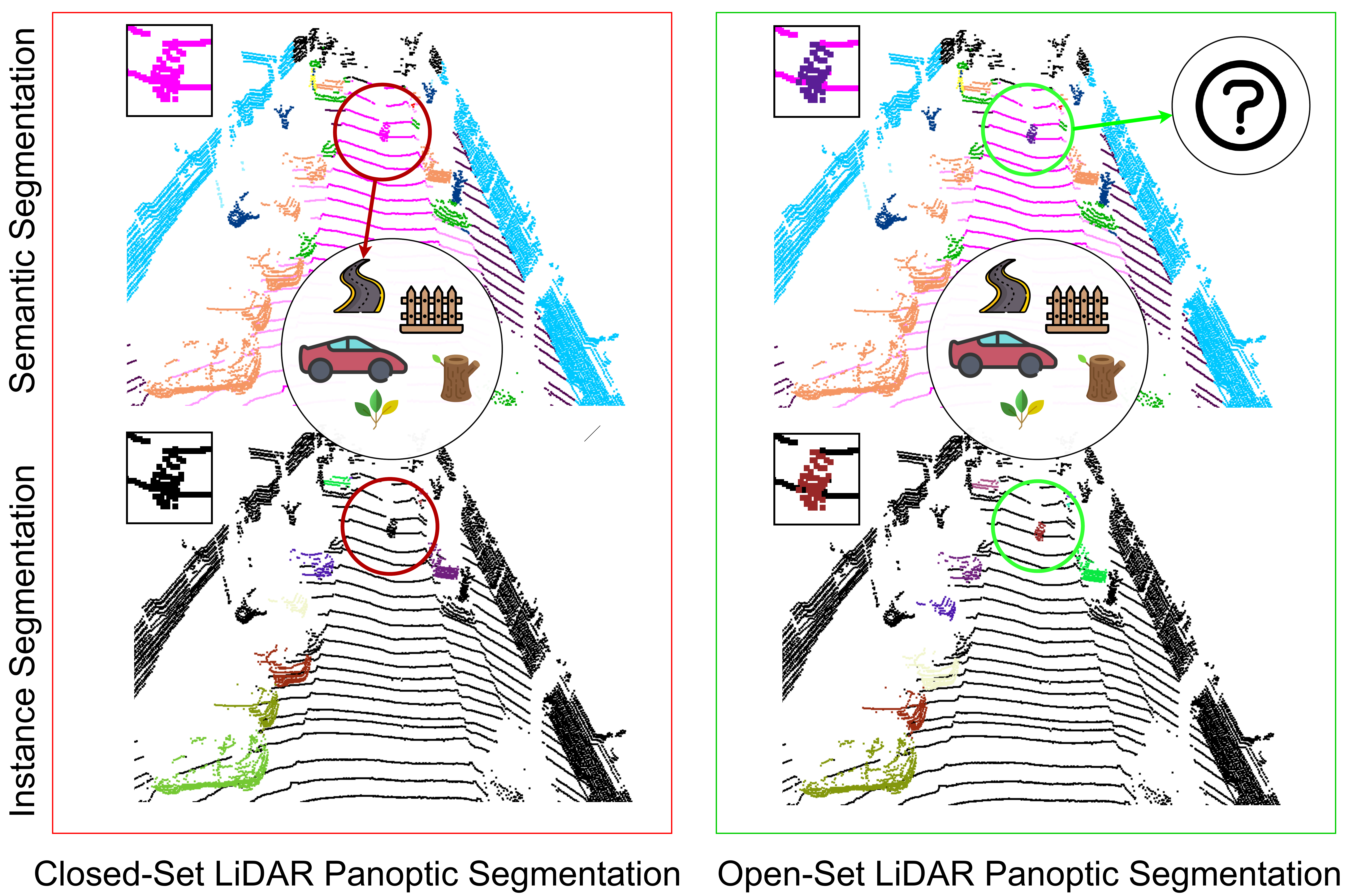}
    \caption{Illustration of the comparison between closed-set and open-set LiDAR panoptic segmentation. In closed-set panoptic segmentation (left), models are limited to predefined categories and misclassify novel objects as known classes, leading to incorrect predictions. In open-set panoptic segmentation (right), models distinguish unknown objects from known categories, enabling reliable scene understanding in open-world environments. \looseness=-1}
    \label{fig:intro}
     \vspace{-0.4cm}
\end{figure}

Standard closed-set models tend to confidently misclassify unknown classes as known ones, leading to inaccurate perception and potential safety risks~\cite{kappeler2024few,gosala2023skyeye,mohan2022perceiving,bevsic2022unsupervised, mohan2024progressive}.
Addressing this challenge requires a shift toward open-set panoptic segmentation, as shown in~\figref{fig:intro}. While this task has been extensively studied in computer vision~\cite{vaze2021open, guo2021conditional, sun2023survey}, its potential for 3D scene understanding remains largely unexplored. One of the first efforts in this direction was OSIS~\cite{wong2020identifying}. It introduced open-set instance segmentation for 3D point clouds by clustering points in a category-agnostic embedding space. This allowed segmentation of both known and unknown objects. However, OSIS relies only on feature-space clustering and does not explicitly distinguish between known and unknown objects. Consequently, it struggles to separate novel instances from background noise or unstructured regions in the scene. Following this, OWL~\cite{chakravarthy2024lidar} extended open-set recognition to panoptic segmentation. It proposed a framework that first classifies points into known stuff, known things, and unknown categories. After classification, it groups unknown and thing instances. However, inspired by~\cite{kong2021opengan}, OWL relies on a fixed "other" category to represent unknowns during training. This limits its ability to generalize and adapt to truly novel objects.

To address the limitations of existing approaches, we propose \net, an uncertainty-guided open-set panoptic segmentation framework for LiDAR point clouds. \net{} explicitly models uncertainty using Dirichlet-based evidential learning, enabling the network to dynamically identify unknown objects at inference time. We introduce a multi-task architecture with three decoders: a semantic segmentation decoder that predicts class labels while quantifying uncertainty, an embedding decoder that learns instance-aware representations, and an instance center decoder that detects object centers. Crucially, we formulate three uncertainty-driven loss functions to reinforce the network’s ability to distinguish between known and unknown regions. Uniform Evidence Loss guides the model to assign high uncertainty to unknown regions, Adaptive Uncertainty Separation Loss ensures a consistent difference in uncertainty estimates between known and unknown objects at a global level, while Contrastive Uncertainty Loss reinforces this separation at the individual sample level. Further, we complement prior work by introducing two novel panoptic open-set evaluation settings based on the KITTI-360~\cite{liao2022kitti} and nuScenes~\cite{fong2022panoptic} datasets.

Our main contributions can be summarized as follows:
\begin{itemize}[topsep=0pt]
    \item We propose \net, an uncertainty-guided open-set panoptic segmentation framework for LiDAR point clouds.  
    \item We design three novel uncertainty-driven loss functions that induce open-set model capabilities. 
    \item Our approach outperforms established open-set LiDAR panoptic segmentation baselines, demonstrating superior capability in distinguishing known and unknown objects.  
    \item We establish new benchmark settings for open-set LiDAR panoptic segmentation on KITTI-360 and Panoptic nuScenes. 
\end{itemize}

%% file: sections/02_related-work.tex
\section{Related Work}\label{sec:related-work}
{\parskip=2pt
\noindent\textit{LiDAR Panoptic Segmentation}: 
LiDAR Panoptic Segmentation methods are typically categorized into single-stage and two-stage approaches. Both approaches rely on a backbone network to extract features from LiDAR point clouds, utilizing different representations such as points~\cite{xiao2025position}, voxels~\cite{zhu2021cylindrical}, range views~\cite{sirohi2021efficientlps}, and bird's-eye views~\cite{zhang2020polarnet, sirohi2023uncertainty}. Two-stage methods~\cite{sirohi2021efficientlps, hurtado2020mopt} fuse instance segmentation outputs from a proposal-based branch with a separate semantic segmentation branch to yield unified panoptic predictions. Single-stage approaches, on the other hand, generate instance predictions using either deterministic methods~\cite{zhou2021panoptic, aygun20214d} or clustering mechanisms that rely on learnable representations~\cite{li2022smac}.
Recent works~\cite{su2023pups, xiao2025position}, inspired by the 2D panoptic segmentation model MaskFormer~\cite{cheng2022masked} predict instance masks and semantic labels in a single pass by employing learnable queries. PUPS~\cite{su2023pups} employs bipartite matching for point-level classification
while MaskRange~\cite{gu2022maskrange} transforms point clouds into a range-view representation to leverage MaskFormer-like architectures.
while P3Former~\cite{xiao2025position} integrate multi-scale voxel features and novel position embeddings.
However, the outlined works are restricted to closed-world scenarios where all categories are observed during training. In this work, we introduce the integration of prototype-embedding association and uncertainty estimation in a single-stage approach to handle open-set scenarios in the LiDAR domain, enhancing segmentation robustness in complex and dynamic environments.}\looseness=-1

{\parskip=2pt
\noindent\textit{Open-Set Learning}: 
Open-set perception has been extensively studied in classification tasks to tackle the challenge of encountering novel classes during inference~\cite{vaze2021open, guo2021conditional, sun2023survey}. To mitigate the limitations of softmax-based classifiers, OpenMax~\cite{bendale2016towards} re-calibrates classification scores to bound the risk associated with unknown classes. An alternative synthetic~\cite{mohan2024syn, sekkat2024amodalsynthdrive} approach, G-OpenMax~\cite{ge2017generative}, employs generative models to synthesize data for unknown classes during training, enhancing the model's capability to identify novel examples during inference. Further, deep generative adversarial networks~\cite{neal2018open, ditria2020opengan} and variational autoencoders (VAEs)~\cite{sun2020conditional, yoshihashi2019classification} have been leveraged to simulate unknown objects during training, reinforcing the model’s ability to distinguish novel classes~\cite{krishnan2018bar,  shi2020multifaceted}.
Recent studies~\cite{malinin2018predictive, charpentier2020posterior} have demonstrated that predictive uncertainty is an effective scoring mechanism to identify objects that do not belong to any known class. For instance, DPN~\cite{sensoy2018evidential} utilizes Dirichlet distributions to capture uncertainty through probabilistic assignments, while Bayesian SSD~\cite{miller2018dropout} uses Monte Carlo Dropout to identify unknown objects during detection. Beyond object recognition, open-world recognition methods~\cite{cen2022open} dynamically label unknowns and incorporate them into subsequent training. In this work, we leverage predictive uncertainty to identify novel classes while proactively shaping the model’s understanding of unknown regions through local and global uncertainty-guided supervision during training, enabling a more structured and adaptive open-set perception.
}

{\parskip=2pt
\noindent\textit{Open-Set Panoptic Segmentation}: 
Compared to its well-established closed-set counterpart, open-set panoptic segmentation remains a challenging task in both images and point clouds. In images, the pioneering model EOPSN~\cite{hwang2021exemplar} groups unlabeled objects across inputs. For LiDAR panoptic segmentation, OSIS~\cite{wong2020identifying} establishes open-set instance segmentation by learning a category-agnostic embedding space. A detection head localizes known objects, while a separately trained embedding head clusters unknown instances using association scores and DBSCAN. Further, OWL~\cite{chakravarthy2024lidar} proposes an open-set learning approach that employs $(K+1)$ classification to capture rare and novel categories, followed by a hierarchical segmentation tree for clustering unknowns. The model is trained on the SemanticKITTI dataset, where some categories are labeled as unknown. Inference is then performed on KITTI-360. However, due to the semantic overlap between the two datasets, OWL is exposed to the same subset of unknown categories during training and inference. We believe that true open-set generalization requires a strict separation between the unknown categories encountered during training and those encountered during inference. Consequently, we propose a new vocabulary split, ensuring that SemanticKITTI classes are excluded from the unknown categories in KITTI-360.
}

%% file: sections/03_methodology.tex
\section{\net~Network Architecture }

\begin{figure*}
    \vspace{0.1em}
    \centering
    \includegraphics[width=\linewidth]{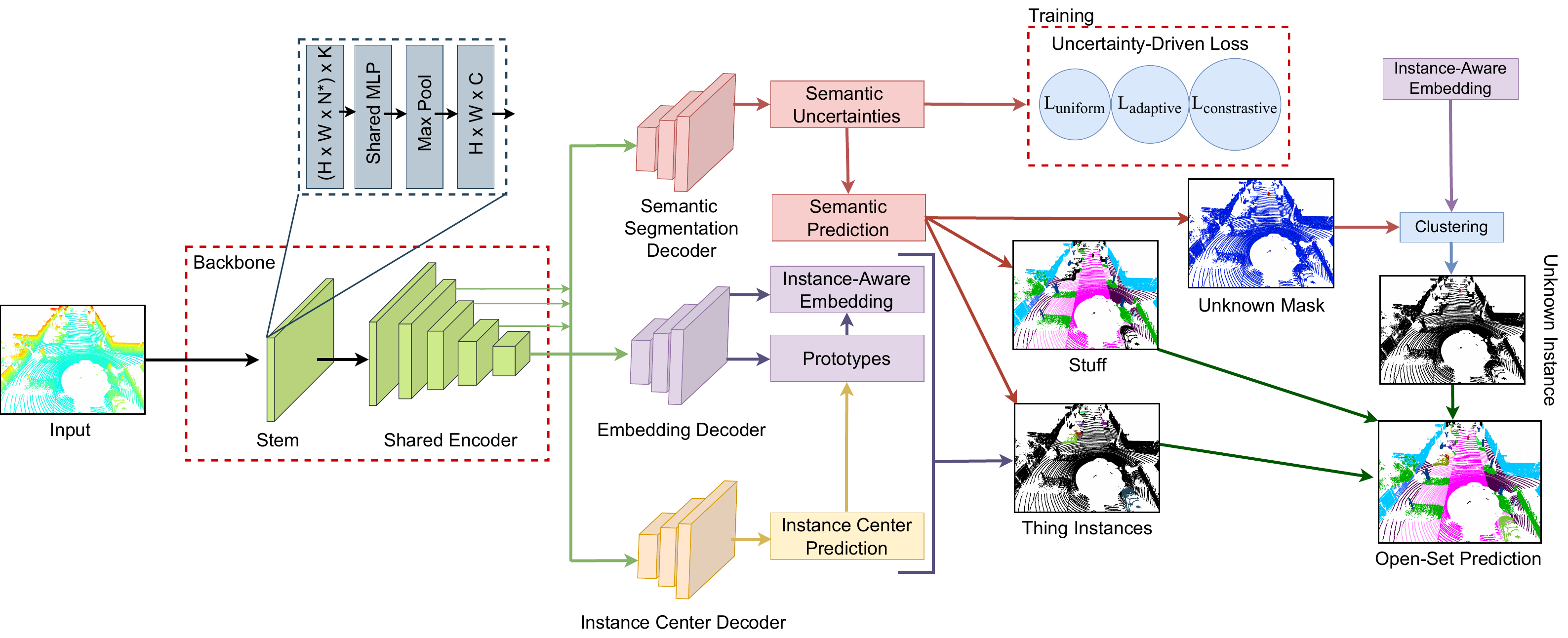}
    \caption{Overview of the \net~architecture for open-set LiDAR panoptic segmentation. The input LiDAR point cloud is processed by a shared backbone, which extracts multi-scale features. Three task-specific decoders refine these features: the semantic segmentation decoder predicts class labels and uncertainty estimates, the embedding decoder generates instance-aware embeddings with prototype association, and the instance center decoder detects object centers. During inference, uncertainty estimates guide the separation of known and unknown regions. Then, known instances are segmented using instance embeddings and object centers, while unknown instances are clustered separately. Our framework is trained with uncertainty-driven loss functions to enforce structured separation between known and unknown objects, leading to robust open-set panoptic segmentation.\looseness=-1}
    \label{fig:network}
     \vspace{-0.4cm}
\end{figure*}

In this section, we introduce our \net~architecture, illustrated in \figref{fig:network}. We begin with an overview of the network before detailing its key components. Our framework is based on a backbone consisting of a stem that encodes the raw point cloud into a fixed-size 2D polar BEV representation, followed by a shared encoder that extracts high-level features. Next, we employ three decoders, each dedicated to a specific task: 1)~a semantic segmentation decoder that generates semantic predictions along with uncertainty estimates, 2)~an embedding decoder that learns instance-aware embeddings and prototypes, and 3)~an instance center decoder that estimates class-agnostic object centers.

During inference, we combine these outputs to yield the final open-set panoptic segmentation. We first use the uncertainty estimates to separate known and unknown regions, labeling high-uncertainty areas as potential unknown objects. In lower-uncertainty (i.e., known) regions, instance-aware embeddings are associated with instance center prototypes to segment \textit{thing} objects. Meanwhile, majority voting in semantic prediction assigns semantic classes to these class-agnostic instances. For high-uncertainty regions, embeddings are clustered to identify unknown objects. During training, we encourage higher uncertainty for unknown regions through a set of specialized losses: \la, \lb, and \lc. By jointly optimizing these losses alongside the core panoptic segmentation objectives, the network is better equipped to distinguish between  known and  unknown objects in LiDAR scenes.

\subsection{Backbone}
Following PolarNet~\cite{zhang2020polarnet}, we first convert the raw point cloud \(\mathbf{P} \in \mathbb{R}^{N \times K}\) into a polar bird's-eye view (BEV) grid of size \(H \times W\), where \(H\) and \(W\) correspond to the radial and angular dimensions, respectively. Each grid cell accumulates points based on their 3D polar coordinates. We then employ a simplified PointNet-style module~\cite{qi2017pointnet}, which applies a series of linear transformations interleaved with batch normalization and ReLU activation. After processing each grid cell, we aggregate features along the vertical \(Z\)-dimension using max pooling, producing a fixed-size representation \(\mathbf{M} \in \mathbb{R}^{H \times W \times C}\), where \(C = 512\) is the feature dimension. We adopt polar rather than Cartesian coordinates to ensure a more balanced distribution of points over different ranges, helping the network focus on closer, denser areas without excessive quantization loss~\cite{zhang2020polarnet}. To capture multi‐scale context, we employ a U‐Net‐style encoder~\cite{ronneberger2015u} that progressively downsamples the BEV representation by a factor of $2\times$, $4\times$, $8\times$, and $16\times$. These feature maps are fed to the decoders, allowing the network to integrate both coarse global context and fine local details in subsequent segmentation and instance‐level tasks.

% Here, $Z$ denotes the number of vertical bins in the voxelized space, while $H$ and $W$ correspond to the radial and angular dimensions of the polar BEV grid. Each grid cell accumulates points based on their discretized polar coordinates, grouping them by radial distance, azimuthal angle, and height. We then employ a simplified PointNet-style module~\cite{qi2017pointnet}, which performs a series of linear transformations interleaved with batch normalization and ReLU. 
% %Specifically, we incorporate a point transformer layer~\cite{zhao2021point} between these MLP blocks to capture local geometric relationships more effectively as shown in~\figref{fig:network}. This attention-based mechanism refines each cell's feature representation, particularly for cluttered or distant regions. 
% After processing each cell, we apply a max-pooling operation along the vertical $Z$-dimension, aggregating features within each BEV grid cell to produce a fixed-size representation $\mathbf{M} \in \mathbb{R}^{H \times W \times C}$, where $C$ is the feature dimension. 
% , yielding multiple resolution scales ($\frac{1}{2}, \frac{1}{4}, \frac{1}{8}, \frac{1}{16}$).

\subsection{Decoders}

We employ three task-specific decoders for semantic segmentation, embedding, and instance center prediction. Each decoder follows a U-Net architecture~\cite{ronneberger2015u} consisting of four upsampling stages. At each stage, feature maps are progressively upsampled and fused with the corresponding encoder outputs at the same resolution. This structure preserves coarse context from deeper layers while reintroducing fine-grained details from earlier encoder stages. While the overall decoding architecture remains consistent, the final layers and feature transformations are tailored to each task. In the following subsections, we provide a detailed explanation of these adaptations.

\subsubsection{Semantic Segmentation Decoder}

The semantic decoder yields class logits for the \( K \) known classes, along with an associated uncertainty, on a three-dimensional grid of size \( Z \times H \times W \).
Here, \(Z \) represents the number of bins in the voxelized 3D space, determining the resolution along the height axis.
Specifically, for each \( (h, w) \) position in the BEV plane, it outputs a prediction of shape \( (Z, K) \), where each vertical level \( z \) contains class logits for all \( K \) known categories. These logits are then reshaped into a 3D voxel representation. During training, each voxel’s ground-truth label is determined by majority voting among the raw LiDAR points that fall within that voxel. Rather than directly applying a softmax function and optimizing with cross-entropy loss, we adopt a Dirichlet-based learning framework~\cite{sensoy2018evidential} to explicitly model uncertainty. The standard cross-entropy loss would force the network to produce deterministic probability distributions, making it overconfident in ambiguous or unknown regions. %This limitation makes it difficult to distinguish between confidently classified voxels and those with high uncertainty due to novel or poorly represented objects.

To address this limitation, the semantic decoder predicts Dirichlet parameters instead of class probabilities. Consequently, this decoder outputs strictly positive parameters \( \alpha_{k} \) via softplus activation, which represents a Dirichlet distribution \( \text{Dir}(\boldsymbol{\alpha}) \) over the class probabilities. We leverage evidence-based learning~\cite{sensoy2018evidential}, where each class’s strength of association is captured by the evidence $e_k = \alpha_k - 1.$ The total uncertainty is then derived as:
\begin{equation}
u = \frac{\sum_{k=1}^{K} \alpha_k}{K},
\end{equation}
which increases when a voxel lacks strong evidence for any known class. 
The network is optimized using the negative expected log-likelihood of the ground-truth class under the Dirichlet distribution:
\begin{equation}
\mathcal{L}_{\text{seg}} = \psi \left( \sum_{k=1}^{K} \alpha_k \right) - \psi(\alpha_y),
\end{equation}
where \( \psi \) is the digamma function, and \( \alpha_y \) corresponds to the voxel’s ground-truth class label \( y \). By learning through the Dirichlet distribution, the model naturally distinguishes known from unknown objects, as voxels with low evidence across all known classes exhibit high uncertainty predictions.

\subsubsection{Embedding Decoder}

Similar to the semantic decoder, our embedding decoder outputs a 3D feature volume of shape \( \boldsymbol{\phi}_{\text{embed}} \in \mathbb{R}^{F \times Z \times H \times W} \). Here, \(F \) is the dimension of the embedding space. %Each \( (h, w) \) cell in the BEV plane generates a \( (Z \times F) \) embedding along the vertical axis, which is then reshaped into a voxel representation. 
As in the semantic branch, voxel-aggregated training labels are assigned via majority voting of same-voxel raw LiDAR points. To facilitate instance-level grouping, we learn prototypes around object centers, which are taken from ground-truth annotations during training. Each prototype $c$ is characterized by a mean \( \boldsymbol{\mu}_c \in \mathbb{R}^{F} \) and variance \( \boldsymbol{\sigma}_c^2 \in \mathbb{R}^{+} \). The voxel embeddings are then compared to these prototypes using a variance-scaled distance metric, producing an association score \( S(v, c) \)~\cite{wong2020identifying} for each voxel–prototype pair:
\begin{equation}
S(v, c) = \exp \left( - \frac{\| \boldsymbol{\phi}^{\text{embed}}_v - \boldsymbol{\mu}_c \|^2}{2 \boldsymbol{\sigma}_c^2} \right),
\end{equation}
where \( \boldsymbol{\phi}^{\text{embed}}_v \) is the embedding of a given voxel $v$. 
During inference, voxels are assigned to the prototype with the highest association score, grouping them into coherent thing instances. We employ a discriminative loss~\cite{de2017semantic} to ensure that embeddings within the same instance form compact clusters while keeping different instances well separated. Additionally, we use a prototype loss~\cite{wong2020identifying} to align the learned prototypes with their corresponding object embeddings. We present the details of these losses in the supplementary material.

\subsubsection{Instance Center Decoder}

The instance center decoder produces a heatmap representing the likelihood of instance centers across the 3D grid of size \( Z \times H \times W \). We assign each voxel a scalar value indicating its probability of being an instance center. Given that object centers may not align precisely with dense LiDAR points in the voxelized representation, we generate the ground-truth center map using a 3D Gaussian kernel centered at each instance’s mass centroid~\cite{aygun20214d}. Specifically, the ground-truth heatmap is defined as:
\begin{equation} 
H_v = \max_i \exp\left(-\frac{\| v - c_i \|^2}{2\sigma^2}\right),
\end{equation}
where \( c_i \) denotes the centroid of instance \( i \) in the voxel space. 
During training, the predicted center heatmap \( \hat{H} \) is supervised using a mean squared error (MSE) loss:
\begin{equation} 
\mathcal{L}_{\text{center}} = \frac{1}{N} \sum_{v} (\hat{H}_v - H_v)^2,
\end{equation}
where \( N \) is the total number of voxels in the grid. The local maxima in the predicted heatmap  \( \hat{H} \) are identified as instance centers which are subsequently used to guide instance grouping in the embedding decoder.

\subsection{Inference}
To achieve open-set panoptic segmentation, we sequentially fuse outputs from the semantic segmentation, embedding, and instance center decoders in three steps. First, unknown regions are identified by thresholding the predicted uncertainty map, where a voxel is classified as unknown if its uncertainty \( u(v) \) exceeds an adaptive threshold, i.e., \( u(v) \geq \mu_u + t \cdot \sigma_u \), where \( \mu_u \) and \( \sigma_u \) are the mean and standard deviation of uncertainties, and \( t \) controls sensitivity to unknown objects.

Next, we perform instance segmentation for known objects by assigning voxels to the nearest predicted center by leveraging the association scores computed in the embedding decoder. We determine the semantic labels through majority voting of the semantic predictions of within each voxel. Finally, following~\cite{wong2020identifying},  we apply DBSCAN~\cite{learndensity} clustering to the embeddings of the embedding decoder to segment distinct instances for unknown voxels.

\subsection{Uncertainty-Driven Losses for Unknown Segmentation} \label{sec:uncertainty_losses}

To ensure the network computes high uncertainty scores for unknown (open-set) regions, we propose three complementary uncertainty losses: \la, \lb, and \lc. These losses are applied to the semantic decoder's output and are computed based on ground-truth unknown labels. Drawing inspiration from evidential deep learning~\cite{malinin2018predictive} and open-set recognition~\cite{winkens2020contrastive}, these losses minimize known class confidence in unknown regions, thereby improving unknown class segmentation at inference time.

\paragraph{Uniform Evidence Loss} 
For each voxel labeled as unknown in the training data (i.e., not belonging to any of the \( K \) known classes), we encourage the Dirichlet parameters to be close to \( \alpha_k \approx 1 \). Equivalently, this implies zero evidence (\( \alpha_k - 1 \approx 0 \)) for all known classes. Concretely, we use a mean-squared error (MSE) penalty:
\begin{equation}
\mathcal{L}_{\text{uniform}} = \| \boldsymbol{\alpha}_{\text{unknown}} - \boldsymbol{1} \|^2,
\end{equation}
where \( \boldsymbol{\alpha}_{\text{unknown}} \) represents the vector of Dirichlet parameters for an unknown voxel, and \( \boldsymbol{1} \) is a vector of ones. This voxel-wise loss ensures that each unknown voxel’s evidence \( \alpha_k - 1 \) remains close to zero. This simple formulation increases the local uncertainty whenever the voxel represents the unknown class, resulting in a flat (maximally uncertain) Dirichlet distribution. However, since this loss is independently applied for each voxel, it does not account for the global distribution of uncertainty in the scene. 

\paragraph{Adaptive Uncertainty Separation Loss} 
While uniform evidence loss enforces high per-voxel uncertainty for unknowns, it does not explicitly ensure a clear separation between the uncertainties of known and unknown voxels. To complement the uniform evidence loss, our adaptive variant uses global statistics to modulate the push for higher uncertainty. We track the mean uncertainty over known (\(\mu_{\text{known}}\)) and unknown (\(\mu_{\text{unknown}}\)) voxels within each batch:
\begin{equation}
\Delta u = (\mu_{\text{unknown}}(u)) - (\mu_{\text{known}}(u)),
\end{equation}
and employ a decaying function of \( \Delta u \) defined as:
\begin{equation}
\mathcal{L}_{\text{adaptive}} = \exp\left(-\frac{\Delta u}{\sigma_{\text{known}}}\right),
\end{equation}
where $\sigma_{\text{known}}$ is the standard deviation of the uncertainty for known voxels. This loss is combined with a training schedule (e.g., an epoch-based weight) to avoid over-penalizing the model once it learns to distinguish known and unknown voxels effectively. This ensures that unknown voxels not only exhibit high uncertainty but are also consistently more uncertain than known voxels. Our loss differs from typical margin-based approaches by considering the distribution of uncertainties on the known classes at each training step. By combining both losses, we enforce high uncertainty at the voxel level (Uniform Evidence Loss) while ensuring a meaningful separation between known and unknown uncertainties at a global level. 

% Consequently, the degree to which we raise unknown uncertainty is adjusted dynamically based on the current separation between known and unknown uncertainty averages, rather than applying a uniform requirement to each voxel.

\paragraph{Contrastive Uncertainty Loss} 
While adaptive uncertainty separation loss ensures a global distinction between known and unknown uncertainties, it does not explicitly enforce per-instance separation. Additionally, Uniform Evidence Loss ensures high uncertainty for unknown regions, but it does so without enforcing a clear relative separation between known and unknown uncertainties. This means that while unknown voxels have high uncertainty, their values may still overlap with certain known voxels, particularly in regions with high intra-class variation. Motivated by contrastive open-set learning~\cite{winkens2020contrastive}, we explicitly compare the uncertainty of known vs. unknown voxels. We sample pairs of (known, unknown) voxels within a batch and enforce $u_{\text{unknown}}~>~u_{\text{known}} + \delta,$ where \( \delta \) is a margin. In practice, we subtract \( \delta \) from the pairwise difference and apply a negative log-sigmoid:
\begin{equation}
\mathcal{L}_{\text{contrastive}} = -\log(\sigma(u_\text{unknown} - u_{\text{known}} - \delta)),
\end{equation}
such that the loss decreases as unknown uncertainty exceeds known uncertainty by at least \( \delta \). This strategy differs from~\cite{winkens2020contrastive} primarily in what is being contrasted: rather than embedding distances, we contrast uncertainty values in our Dirichlet-based segmentation framework. This ensures that any voxel labeled as unknown achieves a distinctly higher uncertainty than any voxel from the known classes.

%% file: sections/04_experiments.tex
\section{Experimental Evaluation}

In this section, we first outline the datasets used for evaluation, followed by the training protocol and benchmarking results. We then conduct ablation studies on different types of uncertainty estimation and the proposed losses, and provide a qualitative analysis. 

\subsection{Datasets}
\label{subsec:dataset}
Following~\cite{chakravarthy2024lidar}, we train on SemanticKITTI~\cite{behley2019semantickitti} and evaluate its open-set performance on KITTI-360~\cite{liao2022kitti}. Both datasets were collected using the same sensor setup in the same city but in different, non-overlapping districts. As KITTI-360 introduces greater variation in unknown classes, its evaluation setup serves as a proxy for LiDAR open-set panoptic segmentation performance. We assess performance on the two vocabulary settings proposed in~\cite{chakravarthy2024lidar}. Vocabulary 1 consists of 6 stuff classes, 3 thing classes, and 1 unknown class, while Vocabulary 2 includes 10 stuff classes, 5 thing classes, and 1 unknown class. 

Additionally, we also report results on SemanticKITTI for completeness. However, due to semantic overlap between the unknown categories of SemanticKITTI and KITTI-360, the standard evaluation setting does not fully measure the impact of truly unseen unknown objects during training. To address this, we introduce a vocabulary referred to as Vocabulary Unseen by modifying Vocabulary 1 for evaluation on KITTI-360, where we exclude unknown semantic classes shared between the two datasets, namely, bicycle, motorcycle, other-vehicle, trunk, pole, traffic-sign, other-structure, other-object, other-ground, and parking. This setup ensures a stricter separation of training and inference unknowns, providing a more comprehensive assessment of open-set generalization.

In addition to SemanticKITTI and KITTI-360, we further evaluate our approach on nuScenes~\cite{fong2022panoptic}, a large-scale dataset for LiDAR panoptic segmentation in urban driving scenarios. This evaluation assess the model’s open-set generalization across a dataset with diverse scene compositions and object distributions. Our defined panoptic open-set vocabulary includes the following thing classes: car, truck, and pedestrian, and stuff classes: road, sidewalk, terrain, manmade, and vegetation. We include bus, construction vehicle, bicycle, motorcycle, and other flat surfaces as unknown categories in the evaluation. Meanwhile, we designate barrier, traffic cone, and trailer as unknown categories during training. In total, our setup consists of three thing classes, five stuff classes, and one unknown class.
\subsection{Training Protocol}
\label{subsec:trainig}
We discretize the 3D space to $[480, 360, 32]$ voxels for SemanticKITTI ($r: 1 \sim \SI{60}{\meter}, z: -3 \sim \SI{3}{\meter}$) and for nuScenes ($r: 0 \sim \SI{50}{\meter}, z: -5 \sim \SI{3}{\meter}$), where $r$ denotes the radial distance from the LiDAR sensor. The network is trained for 60 epochs with a batch size of 4 and a crop size of $[240, 180, 32]$. We use the Adam optimizer with an initial learning rate of $0.01$, with stepwise decay by a factor of 10 at 45 and 55 epochs. We apply instance augmentation along with random point cloud augmentations (flipping, rotation, scaling). We set $t = 3$ for the uncertainty threshold and $F = 32$ for the embedding size. Loss weights are set as follows: $\mathcal{L}_{\text{center}} = 200$, $\mathcal{L}_{\text{uniform}} = 0.1$, $\mathcal{L}_{\text{adaptive}} = 0.1$, and $\mathcal{L}_{\text{contrastive}} = 0.7$, with all other losses weighted at 1.

\begin{table}[]
    \centering
    \caption{Open-set LiDAR panoptic segmentation results on SemanticKITTI and KITTI-360 using vocabularies from~\cite{chakravarthy2024lidar}.}
    \vspace*{-0.2cm}
    \setlength{\tabcolsep}{3pt} 
    \begin{threeparttable}
    \begin{tabular}{lll|ccc|ccc}
        \midrule
        & & &\multicolumn{3}{c}{\textbf{Known}} & \multicolumn{3}{c}{\textbf{Unknown}}  \\ 
        & & \multirow{-2}{*}{Method} & PQ & PQ$^{Th}$ & PQ$^{St}$  & UQ  & Recall & SQ  \\ 
        \midrule

        \multirow{8}{*}{\rotatebox{90}{Vocabulary 1}} &\multirow{4}{*}{\rotatebox{90}{SemKITTI}} & 4D-PLS~\cite{aygun20214d} & 67.8  & 60.0 & 71.7 & 7.8 & 10.8 & 71.9  \\
        & & PolarSeg-Panoptic~\cite{zhou2021panoptic} & 68.6   & 68.2 & 68.8 & 10.2 &14.7  & 69.3 \\

        & & OWL~\cite{chakravarthy2024lidar} & 69.4 & 64.7 & 71.7 &  39.6 & 48.4 & 81.8 \\
        & & \net~(Ours) & 70.5  & 68.9 & 71.3 &  41.9 & 50.6 & 82.3 \\
        \cmidrule{2-9}
        & \multirow{4}{*}{\rotatebox{90}{KITTI-360}} & 4D-PLS~\cite{aygun20214d} & 56.1 &   56.2 & 56.0  & 1.3 &  2.0 &  65.7  \\
        & & PolarSeg-Panoptic~\cite{zhou2021panoptic} & 0.7 &  0.7 & 0.7 & 0.0 &  0.1 &  76.4  \\
        & & OWL~\cite{chakravarthy2024lidar} & 59.4 & 66.2 & 56.0 & 36.3 & 45.1 &  80.5 \\
        & & \net~(Ours) & 60.6  & 67.1 & 57.4 &  38.6 & 47.3 & 80.9 \\
        \midrule
        \midrule
        \multirow{8}{*}{\rotatebox{90}{Vocabulary 2}} &\multirow{4}{*}{\rotatebox{90}{SemKITTI}} & 4D-PLS~\cite{aygun20214d} & 60.2 & 57.9 & 61.4& 16.4 & 22.2 & 73.8  \\
        & & PolarSeg-Panoptic~\cite{zhou2021panoptic} & 58.6  & 56.5 & 55.2 & 14.9 & 20.7 & 72.1 \\
        & & OWL & 61.9  & 62.9 & 61.4 & 49.3 & 57.0 & 86.3  \\
        & & \net~(Ours) & 62.1  & 64.1 & 61.1 &  50.1 & 59.6 & 87.1 \\
        \cmidrule{2-9}
        & \multirow{4}{*}{\rotatebox{90}{KITTI-360}} & 4D-PLS~\cite{aygun20214d}& 45.9  &  47.9 &  44.8 &  3.4 & 4.9 & 69.8\\
        & & PolarSeg-Panoptic~\cite{zhou2021panoptic} &  1.2 &  0.4 &  1.6 &  0.0 & 0.0 &  71.1 \\
        & & OWL & 53.4 & 55.1 & 52.5 & 21.2 &  26.8 & 79.0  \\
        & & \net~(Ours) & 55.0 & 58.3 & 53.4 &  25.9 & 28.5 & 79.7 \\
        \midrule

    \end{tabular}
    Superscripts Th and St refer to \textit{thing} and \textit{stuff} classes, respectively. All metrics are reported in \%.
    \end{threeparttable}
    \label{tab:semkitti}
\end{table}

\begin{table}[]
    \centering
    \caption{Comparison of Panoptic nuScenes using Vocabulary as described in~\secref{subsec:dataset}.}
    \vspace*{-0.2cm}
    \setlength{\tabcolsep}{3pt}
    \begin{threeparttable}
    \begin{tabular}{l|ccc|ccc}
        \midrule
        &\multicolumn{3}{c}{\textbf{Known}} & \multicolumn{3}{c}{\textbf{Unknown}}  \\ 
        \multirow{-2}{*}{Method} & PQ & PQ$^{Th}$ & PQ$^{St}$  & UQ  & Recall & SQ  \\ 
        \midrule
         4D-PLS~\cite{aygun20214d} & 70.7  & 64.3 & 73.9 & 8.3 & 4.8 & 68.3  \\
         PolarSeg-Panoptic~\cite{zhou2021panoptic} & 71.5  & 70.1 & 72.3 & 5.7 &3.7  & 67.2 \\
         OWL~\cite{chakravarthy2024lidar} & 71.6  & 67.2 &\textbf{73.9} &  18.9 & 20.7 & 69.3 \\
         \net~(Ours) & \textbf{72.1}  & \textbf{70.6} & 72.8 &  \textbf{27.3} & \textbf{30.6} & \textbf{71.2} \\
        \midrule

    \end{tabular}
    Superscripts Th and St refer to \textit{thing} and \textit{stuff} classes, respectively. All metrics are reported in \%.
    \end{threeparttable}
    \label{tab:nuscenes}
    \vspace{-.3cm}
\end{table}

\begin{table}[t]
\centering
    \caption{Performance comparison on KITTI-360 with our proposed Vocabulary Unseen as described in \secref{subsec:dataset}.}
    \vspace*{-0.2cm}
    \setlength{\tabcolsep}{3pt}
    \begin{threeparttable}
    \begin{tabular}{l|ccc|ccc}
        \midrule
        &\multicolumn{3}{c}{\textbf{Known}} & \multicolumn{3}{c}{\textbf{Unknown}}  \\ 
        \multirow{-2}{*}{Method} & PQ & PQ$^{Th}$ & PQ$^{St}$  & UQ  & Recall & SQ  \\ 
        \midrule
         4D-PLS~\cite{aygun20214d} & 57.2 & 57.1 & 57.3 & 1.1 & 2.2 & 66.9  \\
         % PolarSeg-Panoptic~\cite{zhou2021panoptic} & 0.9  & 0.9 & 0.9 & 0 &0.1  & 76.2 \\
         OWL~\cite{chakravarthy2024lidar} & 60.7  & 67.5 & 57.3 &  23.1 & 28.6 & 78.3 \\
         \net~(Ours) & \textbf{61.5} & \textbf{68.3} & \textbf{58.1} &  \textbf{35.3} & \textbf{45.3} & \textbf{79.8} \\
        \midrule

    \end{tabular}
    Superscripts Th and St refer to \textit{thing} and \textit{stuff} classes, respectively. All metrics are reported in \%.
    \end{threeparttable}
    \label{tab:unseen_vocab_results}
    \vspace{-.3cm}
\end{table}

\subsection{Benchmarking Results}
\label{subsec:benchmark}

In \tabref{tab:semkitti}, we compare the performance of our \net~with 4D-PLS~\cite{aygun20214d}, PolarSeg-Panoptic~\cite{zhou2021panoptic}, and OWL~\cite{chakravarthy2024lidar} on the SemanticKITTI and KITTI-360 datasets using two different vocabularies. Our evaluation considers multiple metrics,  including Panoptic Quality (PQ)~\cite{chakravarthy2024lidar} for known classes (split into PQ\textsuperscript{Th} for things and PQ\textsuperscript{St} for stuff), as well as Unknown Quality (UQ)~\cite{chakravarthy2024lidar}, recall, and Segmentation Quality (SQ) for unknown regions. Overall, our approach consistently outperforms all baselines, particularly in identifying and segmenting unknown objects. On SemanticKITTI (Vocabulary 1), methods relying solely on K+1 supervision (such as 4D-PLS~\cite{aygun20214d} and PolarSeg-Panoptic~\cite{zhou2021panoptic}) struggle to generalize beyond the known categories. This is evident given the diverse nature of unknown objects in the dataset, ranging from bicycles to traffic signs and tree trunks, resulting in low unknown detection scores.

The problem is even more pronounced in KITTI-360, where the range of unknown objects is significantly larger, further widening the performance gap. In contrast, OWL and our approach, both designed for open-set segmentation, perform significantly better for both known and unknown classes. While OWL and our method both apply clustering to obtain instances, our method achieves a superior performance due to explicitly modeled uncertainty estimation, which divides known and unknown categories. Our uncertainty-driven losses encourage the network to better recognize ambiguous regions, leading to more reliable unknown object segmentation and improved performance. This uncertainty-aware learning strategy not only enhances UQ but also contributes to a boosted segmentation quality for known classes. The combination of instance-aware embedding and prototype association further strengthens our model’s ability to distinguish between known and unknown categories. The strong PQ\textsuperscript{Th} scores across different settings demonstrate our approach’s effectiveness to process thing classes while maintaining comparable performance on stuff classes (PQ\textsuperscript{St}). 
 
We observe a general decline in known PQ across all methods when shifting from Vocabulary 1 to Vocabulary 2. However, our approach maintains a strong advantage in unknown detection. Specifically, our method boosts UQ up to 50.1\% on SemanticKITTI and 25.9\% on KITTI-360, outperforming all tested baselines. Notably, 4D-PLS and PolarSeg-Panoptic also show improved UQ scores under Vocabulary 2, likely due to the reduced number of unknown categories simplifying the task. 

In \tabref{tab:nuscenes}, we further assess our method on nuScenes~\cite{fong2022panoptic}. Our method leverages training unknown labels to guide uncertainty modeling, which outperforms methods based on K+1 supervision alone. Similarly, in \tabref{tab:unseen_vocab_results}, we report performances on KITTI-360 using our defined Vocabulary Unseen. We observe a similar trend, with our model consistently outperforming all baselines for both known and unknown panoptic segmentation. These results underscore the advantages of integrating uncertainty modeling with structured feature learning. Our uncertainty-driven losses enable the network to effectively delineate the boundaries between known and unknown regions, leading to a more robust and balanced segmentation performance across both known and unknown objects.

\subsection{Ablation Study}
\label{subsec:ablation}

In this section, we present two ablation experiments on key components of our network. First, we investigate various uncertainty estimation techniques without supervision to demonstrate that uncertainty estimation is a promising research direction for LiDAR open-set panoptic segmentation. Next, we examine the impact of our proposed losses, which leverage available unknown labels to refine the network's uncertainty estimation.

\begin{table}[t]
\centering
\caption{KITTI-360 performance comparison of annotation-free uncertainty estimation methods.}
\vspace*{-0.2cm}
\begin{tabular}{l|cc|cc}
\midrule
&\multicolumn{2}{c}{\textbf{Vocabulary Unseen}} & \multicolumn{2}{c}{\textbf{Vocabulary 1}}  \\ 
\multirow{-2}{*}{Method} & PQ & UQ  & PQ & UQ  \\ 
\midrule
 Softmax  & 51.7 & 0.5 & 51.3 & 0.8   \\
 DUQ~\cite{van2020uncertainty} & 54.6  & 5.6 & 53.2 & 12.9 \\
SNGP~\cite{liu2020simple} & 57.9 & 14.6 & 56.7 & 22.8 \\
Dirichlet~\cite{sensoy2018evidential} & 60.4 & 27.9 & 59.1 & 31.2 \\
\midrule
\end{tabular}
\label{tab:unc}
\vspace{-.3cm}
\end{table}

\subsubsection{Comparison of Uncertainty Estimation Methods}
In \tabref{tab:unc}, we compare the performance of four uncertainty estimation methods, namely Softmax, DUQ~\cite{van2020uncertainty}, SNGP~\cite{liu2020simple}, and our Dirichlet-based approach implemented in \net. The results are displayed for Vocabulary Unseen and Vocabulary 1. We note that only the uncertainty estimation module is replaced for this experiment in our \net{} architecture. The baseline approach Softmax yields moderate PQ scores (51.7\% and 51.3\%) but almost negligible UQ, clearly indicating its inability to distinguish unknown objects. DUQ improves UQ modestly (5.6\% and 12.9\%), while SNGP further boosts both PQ and UQ, suggesting that uncertainty-aware techniques enhance open set detection. Notably, the Evidential approach achieves the highest performance, with a PQ of 60.4\% and UQ of 27.9\% on Vocabulary 1, and a PQ of 59.1\% and UQ of 31.2\% on Vocabulary Unseen. Although both vocabularies are treated as unseen in this experiment (with Vocabulary Unseen being a subset of Vocabulary 1), the greater diversity among unknown objects in Vocabulary 1 forces the network to produce more pronounced uncertainty estimates, resulting in higher UQ scores, whereas the lower diversity in Vocabulary Unseen leads to milder uncertainty responses. Overall, these results confirm that uncertainty-aware learning significantly enhances open-set segmentation performance, with our Dirichlet-based approach proving especially effective when handling a broader range of unknown objects.

\begin{table}
\centering
\caption{Imapact of uncertainity-driven losses on KITTI-360 performance.}
\vspace*{-0.2cm}
\begin{tabular}{l|cc|cc}
\midrule
&\multicolumn{2}{c}{\textbf{Vocabulary Unseen}} & \multicolumn{2}{c}{\textbf{Vocabulary 1}}  \\ 
\multirow{-2}{*}{Method} & PQ & UQ & PQ & UQ    \\ 
\midrule
 Proto-Unknowns~\cite{wong2020identifying} & 58.4 & 15.1 & 57.1 & 23.2   \\
 \net & 60.4 & 27.9 & 59.1 & 31.2  \\
 + Uniform Evidence & 61.0  & 31.7 & 59.7 & 34.5 \\
 + Adaptive Uncertainty & 61.1 & 32.3 & 59.9 & 35.1  \\
 + Contrastive Uncertainty & 61.5 & 35.3 & 60.6 & 38.6 \\
\midrule
\end{tabular}
\label{tab:loss}
\vspace{-.3cm}
\end{table}

\subsubsection{Impact of Proposed Losses}
Tab.~\ref{tab:loss} presents the impact of our proposed uncertainty losses on the LiDAR open-set panoptic segmentation performance across two vocabulary settings. We first compare two baseline mechanisms for handling unknowns. The Proto-Unknowns baseline, following~\cite{wong2020identifying}, is implemented in our \net{} by removing the uncertainty estimation component and introducing a learnable global constant \( U \), which serves as the score for not associating with any known prototype. In contrast, the \net{} implicitly identifies unknowns through uncertainty estimation. Notably, the Association-Based Unknowns baseline yields a PQ of 58.4\% and 57.1\% with UQ at 15.1\% and 23.2\% on Vocabulary Unseen and Vocabulary 1 respectively. Meanwhile, the \net{} model demonstrates an improved known-unknwon separation (UQ of 27.9\% and 31.2\%), while also achieving improvement in PQ, indicating that leveraging uncertainty cues can generalize to unseen classes even without explicit unknown annotations. However, exposure to unknown samples significantly improves this separation. Incorporating Uniform Evidence Loss increases UQ (+3.8\% on Vocabulary Unseen, +3.3\% on Vocabulary 1), showing that enforcing a flat Dirichlet distribution for unknown voxels effectively raises per-voxel uncertainty. However, PQ remains relatively unchanged, suggesting that while the network becomes more uncertain about unknowns, it does not necessarily refine segmentation quality. Following, introducing Adaptive Uncertainty Separation Loss further enhances UQ (+0.6\% and +0.6\% over Uniform Evidence Loss), confirming that leveraging batch-level statistics ensures an improved distinction between known and unknown uncertainties. Notably, this improvement comes without a significant PQ drop, suggesting that the loss dynamically balances uncertainty separation without overly penalizing confident predictions for known objects. Finally, Contrastive Uncertainty Loss achieves the most substantial gains, boosting UQ to 35.3\% (+7.4\%) and 38.6\% (+7.4\%) across the two vocabularies while also improving PQ. This validates our hypothesis that explicitly enforcing a margin-based separation between known and unknown uncertainties leads to a more structured and meaningful uncertainty landscape. The simultaneous PQ improvement suggests that this contrastive approach also enhances feature representations.

\begin{figure}
    \vspace{0.1em}
    \centering
    \includegraphics[width=0.9\linewidth]{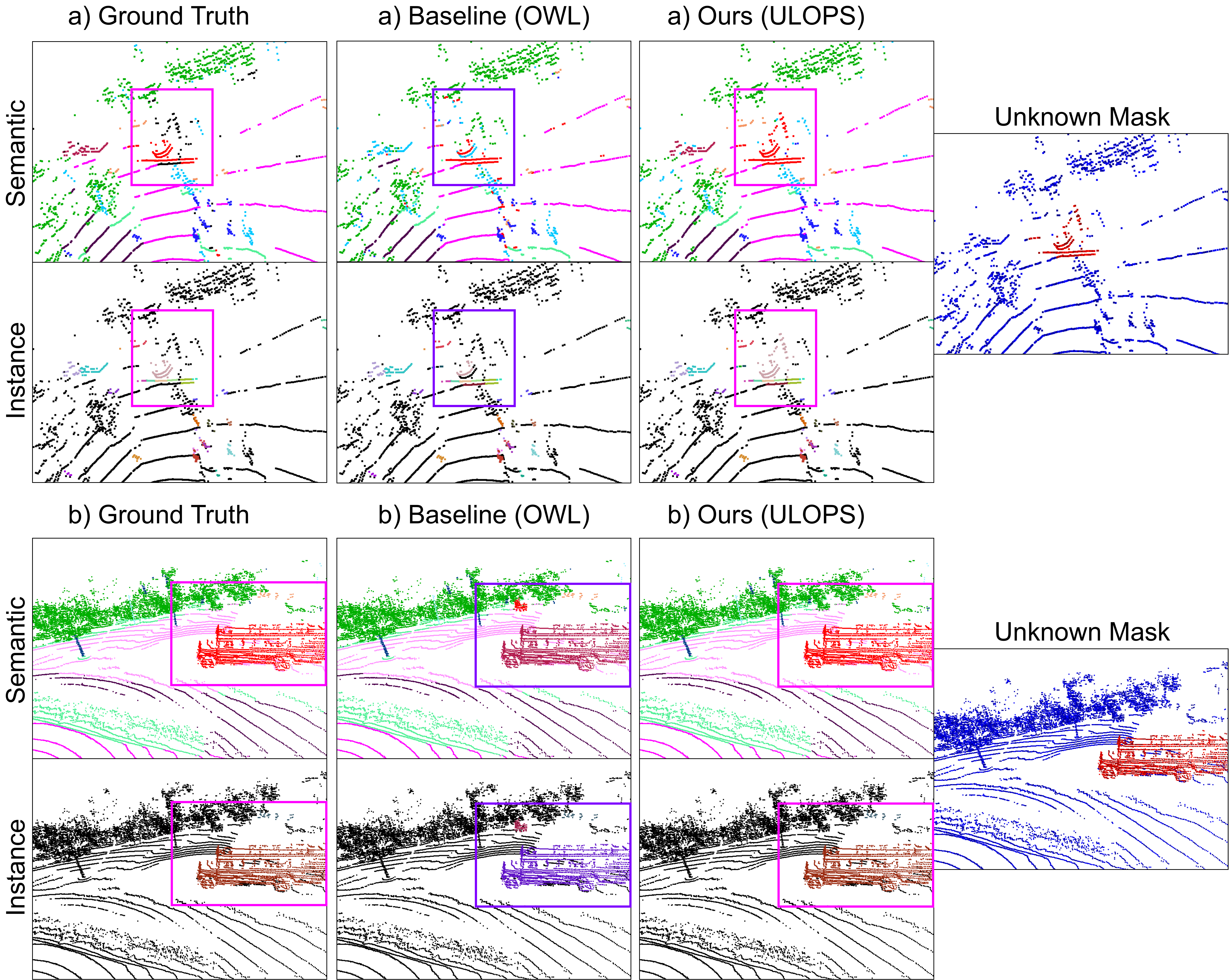}
    \caption{Qualitative results of \net{} compared to the OWL baseline for open-set LiDAR panoptic segmentation. Semantic and instance predictions within the pink box denote correct classifications, while those in the purple box indicate misclassifications. \looseness=-1}
    \label{fig:network}
     \vspace{-0.4cm}
\end{figure}

\subsection{Qualitative Results}
\label{subsec:qualitative}
We qualitatively compare the performance of \net{} with the best-performing baseline, OWL \cite{chakravarthy2024lidar}, as illustrated in Fig. 6. Both models successfully segment barriers in (a). However, OWL struggles to fully segment the construction vehicle, misclassifying parts of it as man-made structures or vegetation. In contrast, \net{} correctly segments the vehicle, demonstrating improved separation between known and unknown objects. Similarly, in (b), OWL misclassifies the unknown object as a truck and falsely detects parts of the vegetation as unknown. \net{} produces a more accurate segmentation, avoiding these errors and providing better differentiation between open-set objects in the scene.

%% file: sections/05_conclusion.tex
\section{Conclusion}
In this work, we introduced \net, an uncertainty-guided open-set LiDAR panoptic segmentation framework. By leveraging Dirichlet-based evidential learning, our method explicitly models uncertainty to identify and separate unknown objects while maintaining segmentation accuracy for known categories. Our multi-task architecture integrates semantic segmentation with uncertainty estimation, instance embedding with prototype association, and instance center detection, enabling robust open-set perception. To enforce a structured distinction between known and unknown objects, we proposed three uncertainty-driven loss functions that improve the model’s ability to detect and segment novel instances. Through extensive experiments on KITTI-360 and nuScenes datasets, we demonstrated that \net{} consistently outperforms existing open-set LiDAR panoptic segmentation methods.

%% file: sections/supplementary.tex
%%%%%%%%%% Merge with supplemental materials %%%%%%%%%%
\clearpage
\renewcommand{\baselinestretch}{1}

\begin{strip}
\begin{center}
\vspace{-5ex}

\textbf{\Large \bf
Open-Set LiDAR Panoptic Segmentation \\ Guided by Uncertainty-Aware Learning
} \\
\vspace{3ex}

\Large{\bf- Supplementary Material -}\\
\vspace{0.4cm}
\normalsize{Rohit Mohan$^{1}$, Julia Hindel$^{1}$,  Daniele Cattaneo$^{1}$, Florian Drews$^{2}$, Claudius Gläser$^{2}$, Abhinav Valada$^{1}$}
\end{center}
\end{strip}

%%%%%%%%%% Merge with supplemental materials %%%%%%%%%%
%%%%%%%%%% Prefix a "S" to all equations, figures, tables and reset the counter %%%%%%%%%%
\setcounter{section}{0}
\setcounter{equation}{0}
\setcounter{figure}{0}
\setcounter{table}{0}
\makeatletter

\renewcommand{\thesection}{S.\arabic{section}}
\renewcommand{\thesubsection}{S.\arabic{section}.\Alph{subsection}}
\renewcommand{\thetable}{S.\arabic{table}}
\renewcommand{\thefigure}{S.\arabic{figure}}

\let\thefootnote\relax\footnote{ $^1$ Department of Computer Science, University of Freiburg, Germany}%
\let\thefootnote\relax\footnote{ $^2$ Bosch Research, Robert Bosch GmbH, Renningen, Germany}%
\normalsize

\normalsize
\input{sections/06_appendix}
\newpage
{\footnotesize
\bibliographystyleS{IEEEtran}
% \bibliographyS{references}
}

%% file: sections/06_appendix.tex
In this supplementary material, we provide additional details on the losses used to train the embedding decoder of ULOPS. We first describe the discriminative loss in~\secref{sec:discriminative}, followed by the prototype loss in~\secref{sec:prototype}.

\section{Discriminative Loss}
\label{sec:discriminative}

To enforce compact and separable instance embeddings, we adopt a discriminative loss composed of an intra-cluster pull term and an inter-cluster push term. These components jointly encourage embeddings belonging to the same instance to cluster tightly around a prototype, while ensuring that different instances remain well-separated in the embedding space.

The pull loss minimizes the distance between each voxel embedding and the prototype of its corresponding instance:
\begin{equation}
L_{\text{pull}} = \frac{1}{\sum_{c=1}^{N} |C_c|} \sum_{c=1}^{N} \sum_{v \in C_c} \left\| \boldsymbol{\phi}_{\text{embed}}(v) - \boldsymbol{\mu}_c \right\|^2,
\end{equation}
where \( N \) is the number of instances, \( C_c \) denotes the set of voxels belonging to instance \( c \), \( \boldsymbol{\phi}_{\text{embed}}(v) \in \mathbb{R}^d \) is the embedding vector of voxel \( v \), and \( \boldsymbol{\mu}_c \in \mathbb{R}^d \) is the prototype (mean embedding) of instance \( c \).

The push loss enforces separation between prototypes of different instances by penalizing those that are closer than a margin \( \Delta \):
\begin{equation}
L_{\text{push}} = \frac{1}{N(N-1)} \sum_{c \neq c'} \max\left(0, \Delta - \left\| \boldsymbol{\mu}_c - \boldsymbol{\mu}_{c'} \right\|^2\right),
\end{equation}
where \( \Delta > 0 \) is a margin hyperparameter to prevent prototype collapse.

\section{Prototype Loss}
\label{sec:prototype}

To reinforce prototype consistency, we employ a prototype alignment loss, which encourages the learned prototype \( \boldsymbol{\mu}_c \) of each instance to match the average of the voxel embeddings assigned to that instance:

\begin{equation}
L_{\text{proto}} = \frac{1}{N} \sum_{c=1}^{N} \left\| \boldsymbol{\mu}_c - \frac{1}{|C_c|} \sum_{v \in C_c} \boldsymbol{\Phi}_{\text{embed}}(v) \right\|^2.
\end{equation}

This loss stabilizes prototype updates and complements the pull and push terms by explicitly anchoring prototypes to their corresponding instance embeddings.

\section{Embedding Loss}
\label{sec:embedding_loss}

The total embedding loss is a weighted combination of the components introduced in~\secref{sec:discriminative} and~\secref{sec:prototype}:
\begin{equation}
L_{\text{embed}} = \lambda_{\text{pull}} L_{\text{pull}} + \lambda_{\text{push}} L_{\text{push}} + \lambda_{\text{proto}} L_{\text{proto}},
\end{equation}
where \( \lambda_{\text{pull}}, \lambda_{\text{push}}, \lambda_{\text{proto}} \in \mathbb{R}_{\geq 0} \) are hyperparameters controlling the relative importance of each term. In our experiments, we set \( \lambda_{\text{pull}} = 1 \), \( \lambda_{\text{push}} = 1 \), and \( \lambda_{\text{proto}} = 0.001 \).

By jointly optimizing these objectives, the embedding decoder learns a representation space where embeddings of voxels belonging to the same instance form compact clusters around their prototype, while remaining well-separated from other instances.
 